\documentclass[preprint,12pt]{elsarticle}
\usepackage{epstopdf}
\usepackage{lineno}

\usepackage{lscape}
\usepackage{filecontents}
\usepackage{adjustbox}
\usepackage{url}

\usepackage[font=small,labelfont=bf]{caption}
\usepackage{multirow}
\usepackage{hyperref}
\usepackage[english]{babel}
\usepackage{algorithm,algorithmic}

\usepackage{graphicx}
\usepackage{colortbl}
\usepackage{color}
\usepackage[numbers]{natbib}
\usepackage{chngcntr}
\usepackage{longtable}
\usepackage{adjustbox}
\usepackage{rotating}
\usepackage{subfig}

\usepackage{amsmath,amsfonts,amssymb,amsthm}

\theoremstyle{plain}
\theoremstyle{Definition}
\newtheorem{mydef}{Definition}

\newtheorem{theorem1}{Theorem}

\begin{document}


\begin{frontmatter}
\title{Modified online Newton step based on element wise multiplication }

\author[1,2]{Charanjeet}
\author[1]{Anuj Sharma}

\address[1]{Department of Computer Science and Applications, Panjab University, Chandigarh, India}
\address[2]{Centre for Advanced Study of Mathematics, Panjab University, Chandigarh, India}
\ead[url]{https://sites.google.com/view/anujsharma/}
\begin{abstract}
The second order method as Newton step is a suitable technique in Online Learning to guarantee regret bound. The large data is a challenge in Newton method to store second order matrices as hessian. In this paper, we have proposed a modified online Newton step that store first and second order matrices of dimension $m$ (classes) by $d$ (features). we have used element wise arithmetic operation to retain matrices size same. The modified second order matrix size results in faster computations. Also, the mistake rate is at par with respect to popular methods in literature. The experimental outcome indicates that proposed method could be helpful to handle large multi class datasets in common desktop machines using second order method as Newton step.
\end{abstract}
\begin{keyword}
machine learning, online learning, online Newton step, element wise multiplication, first and second order online learning


\end{keyword}

\end{frontmatter}



\section{Introduction}
Machine Learning (ML) plays decisive role in emerging Artificial Intelligence (AI) applications and its data analysis. ML is a core sub-field of AI; it results in self-learning platform to machines, without being explicitly programmed. These algorithms are enabled to learn, grow, change and develop themselves. Conventional machine learning (batch based) is different from online learning paradigm as they require whole data at the time of training and these techniques are suffering from expensive re-training cost when dealing with a new instance coming. In big data era, conventional learning algorithms are more restrictive when data is huge and complex for real-world applications \cite{Chauhan2018}. Different from batch-based learning; Online Learning (OL) includes important learning algorithms which are developed to learn models incrementally from data in a sequential manner \cite{JMLR:v15:hoi14a},\cite{Onlinesurvey}. OL algorithms overcome the limitation of batch-based learning as OL instantly update the model when new instance arrives. Most of the OL algorithms are inspired from online convex optimization.
An online convex optimization consists of a convex set $S$, and a convex payoff function $l_t$. At step $t$, the online convex optimization choose a weight vector $w_t \in S$ then it computes a loss $l_t(w_t)$, which is based on the convex payoff function $l_t(.)$ defined over a convex set $S$ \cite{DBLP:journals/corr/abs-1811-09955}. The objective of the online learning algorithm to attain a minimum regret bound, where regret is described as:
\begin{equation}
R_t= \sum_{t=1}^{T}l_t(w_t)-\inf_{w^* \in S} \sum_{t=1}^{T}l_t(w^*),
\end{equation}
where $w^*$ is the minimum value of the payoff function as $\sum\limits_{t=1}^{T} l_t(w)$ over the convex set $S$. \\
Based upon the optimization techniques, OL algorithms categorized in the following way: (i) first-order OL where only first-order feature information is used \cite{PA2006},\cite{Zinkevich:2003:OCP:3041838.3041955},\cite{DUOL2011},\cite{NHERD2010}  (ii) second order OL where not only first-order but also second order feature information used in training \cite{SOP2005},\cite{Lin:2007:TRN:1273496.1273567}, \cite{CW2008},\cite{SOOAL}.
The first OL algorithm is Perceptron \cite{Percept1957} based upon the first order derivation of the cost function, which updates the model by adding and subtracting the loss of the misclassified instance with a fixed weight. Recently, various methods were proposed based upon the first order OL by maximizing the margin value. One developed Relaxed Online Margin Algorithm (ROMMA) \cite{ROMMA2002} is an incremental approach based upon the maximum margin. ROMMA used the first-order derivation of the linear function for classification. Another approach is Passive Aggressive (PA) \cite{PA2006} that classified the current instance based upon the loss function, when the loss is zero updation is passive otherwise updation is aggressive. By analyzing performance of the first-order based OL techniques, we observe that large margin method are out-performed the perceptron algorithm. \\
In recent year, second order methods gain popularity as they overcome the limitations of first-order method by exploiting the second order information. The well known second order method is Second-Order Perceptron (SOP) \cite{SOP2005}, which is the variant of whitened perceptron algorithm. In this work the author defined the interaction between eigenvalue of the correlation matrix of the data and the target vector. There are various other large margin second order online learning methods are proposed, such as Confidence-Weighted (CW) \cite{CW2008} learning defined over the notation of confidence parameter. The less confident parameters are updated more aggressively than confident one. The limitation of CW, its aggressive hard margin strategy in noisy data. To overcome this problem, researchers proposed the another variant of CW learning based upon the adaptive regularization of each training instance. In general, the performance and convergence rate of second-order OL is better than the first-order algorithms.  \\
The Online Newton Step (ONS) \cite{ONS2007} is analogue of the Newton-Raphson method \cite{NEWTON}. It moves in the direction of the vector which inverse Hessian multiplied by the gradient. In ONS case, the direction is $H_t^{-1} \nabla_t$ and the matrix $A_t$ is Hessian. Second-order algorithms such as ONS lead to lower regret bound as compared to first-order methods. 
Our main contribution in the proposed work is implementation of second-order algorithm analogue to the Online Newton step. We have kept the dimension of first order and second order derivatives same as objective function using element wise multiplication and inclusion of unit vectors or multiplication with feature vector. Therefore, we are able to reduce the number of iterations required in the previous used ONS. The main idea behind the change in dimension of matrix is to reduce the calculation time of learning system. 
Our first order and second order matrices as $b$ and $A$, do not share same definition as gradient and hessian matrices. Therefore, we have assigned names as matrices $b$ and $A$, differently from gradient and hessian notations, as used in literature work. The highlights of the study are:
\begin{itemize}
\item A suitable second order online learning method using Newton step.
\item First and second order matrices are of order $m \times d$ instead of $d \times d$ as used in literature work.
\item The promising online computation time cost as compared to previous work and of complexity $\mathrm{O}(md)$.
 \item Analysis of regret bound of proposed method.
\end{itemize}
The rest of the paper is organized as follows. The Section 2 presents the notations used in this paper and the role of variables in convex optimization problems. In section 3, we introduce the proposed Online Newton Step for multiclass with the new dimensions of the matrices. In section 4, we present the theoretical analysis and regret bound of the proposed method.  In Section 5, summarized the experiments based on the proposed method with the benchmarks datasets. The section 6 concludes the paper with finding.

\section{Preliminaries}
In this section, we present the notations used in this paper, and formally define the problem of second order convex optimization.
\subsection{Notations}

\begin{table}[htb]
\begin{center}
\caption {Symbol descriptions} \label{tab:tab1}

\begin{tabular}{|l|l|}
\hline
\textbf{ Notations  }                                                                       & \textbf{Definitions}  \\ \hline \hline
$\mathbf{F}$                                              & Convex feasible set        \\
$\mathbf{X}$ &  $\mathbf{  x_1;x_2;....;x_t  }$          \\
$f$ & $\underset{w \in \mathbb{R}^{d \times m}}{\mathrm{min}}
f(w)=\log(1+exp(-y_t <x_t w^T>)+ \lambda ||w||_1$ \\
$l_t$                                 & Convex loss function of $t^{th}$ interation      \\
$\nabla(f)$                                              &  Gradient of the loss function     \\
$\nabla^2(f)$                                                    & Second order derivative of the loss function     \\
$\mathbf{Y_t}$                                             & $y_1;y_2;.....y_t$      \\
$||.||$                                                    & Euclidean norm \\
$R_G(T)$                                  & Regret bound        \\
     $m$                                          & number of classes in data set     \\
$d$                                                & number of feature in each vector $x_t$        \\
$A$   & second-order derivative of $l_t$ \\
$g$   & first order derivative of $l_t$ \\
$A^T$ & Transpose of matrix $A$ \\
$\bigodot$ & $A.^*b$ element-wise multiplication \\
$k$   & $\frac{\nabla_t}{\nabla_t^2}$ \\ \hline
\end{tabular}

\end{center}
\end{table}
In Table \ref{tab:tab1}, we present the main symbols used in this paper. We have used bold upper case character for matrix (e.g $\mathbf{(A)}$)
representation and $\mathbf{A^{-1}}$ and $\mathbf{A^T}$ denote the inverse and transpose of matrix respectively, small bold case character
represent the vector ($\mathbf{x}$). $||x||$ used for Euclidean norm of a vector that is dual to itself. $\nabla$ denote the first order derivative and
$\nabla^2$ denote the second order derivatives of the objective function.\\
In our case, the dimension of the second order derivative is $m \times d$, where $m$ is the number of classes in respective data set and $d$ is the number of features in current vector $\mathbf{x_t}$.
\subsection{Second order online convex optimization}
Many online learning algorithms formulated as an online convex optimization task. The main aim of second order methods is to exploits the second
order information to speed up the convergence of the optimization. The popular second order online algorithm is online Newton step. Like ONS, second order methods update the model in sequence of consecutive rounds. At each round $t$, the online learner pick a data vector $x_t$ from a convex set $F$ such that $x_t \in X$. After the prediction is calculated on vector $x_t$, a convex loss function $l_t$ is revealed, then the online learner suffer from an instantaneous loss $l_t(x)$. In second order online convex optimization, we assume that the sequence of loss function $l_1,l_2,....,l_T : X\rightarrow \mathbb{R}$ are fixed in advance. The main goal of online learner is to minimize the regret. In second order learning, the learner has access to the gradient as well as second order derivative of the loss function at any point in the convex set $F$.
\section{Modified Online Newton Step using element wise multiplication}
In online learning, Newton method is second order optimization algorithm, which iteratively updates the weight vector $w \in \mathrm{R}^{d \times m}$ in a sequential manner of an objective function $f$ by computing direction $d_t$ and update weight vector:
\begin{equation}
w_{t+1} =  w_t - \eta_t . d_t,
\end{equation}
where $\eta_t$ is a step size and $d_t$ is search direction. The benefit of using Newton method is that $f$ can be locally approximated around each $w_t$, upto second order, by the quadratic:
\begin{equation}
f(w_t+d_t) \approx q_{w_t}(d_t) \equiv f(w_t)+ \nabla f(w_t)^T d_t + \frac{1}{2} d_t^T \textbf{H} d_t,
\end{equation}
where $\textbf{H}= H(w_t)$ is the Hessian matrix of $f$ at $w_t$. To move in a good direction to, reduce to minimize this quadratic ($q(w_t)$) with respect to $d_t$, to find a descent direction, we use a Newton algorithm. Computing and storing the full Hessian matrix ($H$) takes $\mathrm{O}(d^2)$ memory, which is infeasible for high-dimensional functions such as the loss functions of neural nets with large numbers of features.\\
The proposed work in this paper is an attempt to overcome the challenges of Hessian complexity as $\mathrm{O}(d^2)$.
The goal of the present work is to facilitate research an online learning optimization for large-scale data. We address a logistic minimization problem defined as:
\begin{equation}
\label{logloss}
\underset{w \in \mathbb{R}^{d \times m}}{\mathrm{min}} f(w)=\log(1+exp(-y_t <x_t w^T>)+ \lambda ||w||_1,
\end{equation}
where $w \in \mathbb{R}^{d \times m}$ represent the model parameter and $T$ define the number of instances $(x_t, y_t)$. $L(<w_t, x_t>, y_t)$ is the loss function.\\
The proposed algorithm is a online variant of the well know Newton method \cite{Lin:2007:TRN:1273496.1273567} for finding the roots. In online learning, \cite{Onlinesurvey} at time $t$, the learner pick one instance $x_t$ from the environment and then make a prediction of it's class label $\hat{y}=sign(f(x_t))$, where $f$ is a prediction function that maps the feature vector to a real valued classification score. After the prediction, the true class label is revealed from the environment and update the classifier, when the learner makes a mistake ($\hat{y_t} \neq y_t$ ). \\
In this paper, we propose a new modified Newton step in online environment, which keeps the dimension of second order matrix as well as first order matrix same. The previous work on second order methods store the full information of second order derivative of the objective function, which results in more time to train the data when data is large. The new algorithm store the matrices $A$ and $b$, the dimension of $A$ and $b$ are $m \times d$, where $m$ is the number of classes and $d$ is the number of features in each instance.
\begin{algorithm}[h]	
	\caption{ONS using element wise multiplication}\label{algo1}
	\begin{algorithmic}[1]
	\STATE \textbf{Input:} learning rate $\eta_t$ ; regularization parameter $\lambda$
		\STATE \textbf{Initialize}: $w_1=0$
		\FOR {i=1 to T}
		\STATE Receive $x_t \in X$;
		\STATE Compute $F_t= -y_t <x_t, w_t^T>$
		\STATE Make a prediction $\hat{y}_t= \mathtt{sign}(F_t)$;
		\STATE Compute Loss $l_t=\texttt{max}(0, 1-F_t)$
		\IF {$l_t > 0$}
		\STATE Compute:$f(w_t)=log(1+exp(-y_t <x_t w_t^T>)+ \lambda ||w_t||_1)$
		\STATE Compute b= $\nabla f(w_t). y_t \bigodot x_t + \lambda w_t$
		\STATE Compute A=$\nabla^2 f(w_t) + \lambda$
		\STATE Compute step size $\eta_t=\frac{1}{2 \sqrt{t}}$
		\STATE Compute the Newton direction: $d_t = A^{-1} \bigodot b $
		\STATE update the classifier :$w_{t+1}= w_t - \eta_t d_t  $
        \ENDIF
        \ENDFOR
	\end{algorithmic}
\end{algorithm}
The proposed algorithm ONS consist of two parts. 1) the size reduction of the second order matrix with respect to dimension. 2) The present rule of updating the classifier $w$. we discuss each part in detail follows.\\
In the algorithm \ref{algo1}, at $t^{th}$ round, the true label of $y_t$ of $x_t$ is revealed, we will update the classifier to make sure that it suffer a small loss on the $t^{th}$ instance and the new classifier $w_t$ is not far away from the previous classifier. Formally, we want to update the classifier by minimizing the objective function that is defined in equation \ref{logloss}. $l_t=max(0,1-y_tw_t^Tx_t)$ is the hinge loss function used.\\
When $l_t > 0$, we solve the minimization problem \ref{logloss} in the following steps:
\begin{itemize}
\item Compute the $\nabla f(w_t)$ and $\nabla^2 f(w_t)$ of equation \ref{logloss}
\begin{equation}
b=\nabla f(w_t)= y_t \bigodot x_t (f(w_t) \bigodot( 1-f(w_t))+ \lambda w_t
\end{equation}
\begin{equation}
A=\nabla^2 f(w_t)= x_t^2 b + \lambda
\end{equation}
\item Update the classifier parameter
\begin{equation}
w_{t+1}=w_t- \eta_t A^{-1} \bigodot b
\end{equation}
\end{itemize}
In the first step, we compute the matrix $b \in \mathrm{R}^{d \times m}$, which is equivalent to first order derivative of the logistic loss function $L_t$ with regularization  parameter $L1$ regularization. 
In order to meet matrices multiplication criteria, we have multiplication with unit vector to fullfill the element wise multiplication criteria. This further help in strengthening our technique and results in low mistake rate.
In the second step, We update the classifier and the updated equation is based upon Newton method for finding the root. Here we update weight matrix $w \in \mathrm{R}^{d \times m}$ and perform element wise $\bigodot$ multiplication. using element wise multiplication, we maintain the first order as well as second order derivative matrices dimensions as $m \times d$. The source code in Matlab for algorithm \ref{algo1} given in Appendix 1.\\
The modify newton step algorithm is straight forward to implement, and running time is $\mathcal{O}(md)$ per iteration give the second order matrix. In this algorithm we use the step size $\eta_t= \frac{0.5}{\sqrt{t}}$.
\subsection{Theoretical Analysis of modify Newton Step}
As discussed by Zinkevich \cite{Zinkevich:2003:OCP:3041838.3041955}, we have follow the same definition 1. Our proof is different from Zinkevich, as we have used $\frac{\nabla_t}{\nabla_t^2}$ instead of $\nabla_t$ only.
\begin{mydef}
\label{mydef1}
A feasible set $F \in \mathrm{R}^n$ and a loss sequence $\{l_1,l_2,...\}$ where $l_t: F \rightarrow \mathrm{R}$ is a convex function in online convex optimization problem. \\
A vector $x_t \in F$, at step $t$, it receives the loss function $l_t$.
\end{mydef}
\begin{mydef}
	\label{mydef2}
In an algorithm $A$, a convex problem for $F$ and $\{l_1,l_2,...\}$, if $\{w_1,w_2,...\}$ are selected vector by $A$, until time $T$, the loss of $A$ is:
\begin{equation*}
L_A(T)= \sum_{t=1}^{T}l_t(w_t)
\end{equation*}
static feasible solution loss for $w \in F$ is
\begin{equation*}
L_w(T)= \sum_{t=1}^{T} l_t(w)
\end{equation*}
Algorithm $A$ regret is
\begin{equation}
R_A(T)=L_A(T)- \underset{w \in F}{min} L_w(T)
\end{equation}
\end{mydef}
 We are specifying some parameters in order to state results in respect of regret of algorithm bounded.
\begin{equation*}
\begin{array}{lll}
\hspace{0.3cm}||F||= \underset{w,w^* \in F}{\mathrm{max}} \ d(w,w^*) \\
||k||= \underset{w \in F , t \in {1,2,...,T}}{\mathrm{sup}} || k_t(w)||,
\end{array}
\end{equation*}
where $d(w,w^*)=||w-w^*||$, here is the first results derived in this paper.
\begin{theorem1}
\label{theorem2}
if $\eta_t= 0.5/ \sqrt{t}$, the regret of the modified Newton step is:
\begin{equation*}
R_G(T) \leq (||F||^2 + ||k||^2) \sqrt{T}
\end{equation*}
Therefore, lim $sup_{T \rightarrow \infty} R_G(T)/T \leq 0$
\end{theorem1}
\begin{proof}
For all $t$, there exist a $k_t \in \mathrm{R}^n$ such that $l_t(w)=k_t \cdot w$ for all $w$. we compute $w_1,w_2,...$ by running the algorithm and begin with arbitrary $l_1,l_2,...$ We define $k_t=\nabla l_t(w_t)$. For all $w$, $l_t(w)=k_t \cdot w$ to change $l_t$, the algorithm behavior is same as:
\begin{equation*}
l_t(w) \geq (l_t(w_t)) \cdot (w-w_t)+l_t(w_t)
\end{equation*}
Therefore, for all $w^* \in F: l_t(w^*) \geq k_t \cdot (w^*-w_t)+ l_t(w_t)$. Thus:
\begin{equation*}
\begin{array}{lll}
l_t(w_t)-l_t(w^*) \leq l_t(w_t)-(k_t \cdot (w^*-w_t)+ l_t(w_t)) \\
\hspace{2.8cm} \leq k_t \cdot w_t - k_t \cdot w^*
\end{array}
\end{equation*}
This regret is least with respect to modified sequence of functions. Now, for $w^*$ on round $t$,we find regret bound as,
\begin{equation*}
\begin{array}{l}
w_{t+1}=w_t - \eta_t \frac{\nabla_t}{\nabla_t^2} = w_t- \eta_t k_t \hspace{2cm} \texttt{where }  k_t= \frac{\nabla_t}{\nabla_t^2} \\
w_{t+1}-w^*=w_t-w^*-\eta_t k_t \\
(w_{t+1}-w^*)^2=((w_t-w^*)-\eta_t k_t)^2 \\
\hspace{2.4cm} =(w_t-w^*)^2-2(w_t-w^*) \eta_t k_t + \eta_t^2 ||k_t||^2
\end{array}
 \end{equation*}
Also, $||k_t|| \leq ||k||$
 \begin{equation*}
 (w_{t+1}-w^*)^2 \leq \frac{1}{2 \eta_t}((w_t-w^*)^2-(w_t-w^*)^2))+ \frac{\eta_t}{2} ||k||^2
 \end{equation*}
Now, by summing we get,
\begin{equation*}
\begin{array}{ll}
 R_G(T)= \sum_{t=1}^{T}(w_t-w^*)\cdot k_t \\
 \hspace{1.2cm}\leq \sum_{t=1}^{T} \frac{1}{2 \eta_t}((w_t-w^*)^2-(w_{t+1}-w^*)^2) + \frac{\eta_t}{2} ||k||^2 \\
 \hspace{1.2cm} \leq \frac{1}{2 \eta_1}(w_1-w^*)^2-\frac{1}{2 \eta_T}(w_{T+1}-w^*)^2+\frac{1}{2} \sum_{t=2}^{T}(\frac{1}{\eta_t}-\frac{1}{\eta_{t-1}})(w_t-w^*)^2+ \frac{||k||^2}{2} \sum_{t=1}^{T} \eta_t \\
 \hspace{1.2cm}  \leq ||F||^2(\frac{1}{2 \eta_1}+ \frac{1}{2} \sum_{t=2}^{T}(\frac{1}{\eta_t}-\frac{1}{\eta_{t-1}}))+ \frac{||k||^2}{2} \sum_{t=1}^{T} \eta_t \\
 \hspace{1.2cm} \leq ||F||^2 \frac{1}{2 \eta_T} + \frac{||k||^2}{2} \sum_{t=1}^{T} \eta_t
   \end{array}
\end{equation*}
Now, if we define $\eta_t= \frac{1}{2 \sqrt{t}}$,
\begin{equation*}
\begin{array}{l}
\sum_{t=1}^{T} \eta_t= \sum_{t=1}^{T} \frac{1}{2 \sqrt{\eta_t}} \\
\hspace{1cm} \leq 1+\frac{1}{2} \int_{t=1}^T 2 \sqrt{t} \\
\hspace{1cm} \leq 1+(\sqrt{T}-1) \\
\hspace{1cm} \leq \sqrt{T}
\end{array}
\end{equation*}
After putting the value of $\sum_{t=1}^{T} \eta_t$, we get regret bound as,
\begin{equation*}
R_G(T) \leq (||F||^2 + ||k||^2) \sqrt{T}
\end{equation*}

\end{proof}
\section{Experiments}
Experiments of the proposed method were performed on multi-class classification datasets, i.e satimage, mnist, acoustic and protein. All the datesets are downloaded from LIBSVM website. We selected the four mutli-class classification datasets and  table \ref{Tab:1} shows the statistics of the datasets used in experimentations. All experiments are performed in MATLAB environment on a single computer with 3.4 GHz Intel core i7 processor and 8 GB RAM running under window 10 operating system.\\
We have compared our proposed modified ONS algorithm against the two following baselines:
\begin{itemize}
\item The modified ONS change the dimension of second order as well as first order matrices which is $A$ and $b$ from $d \times d $ to $m \times d$.
\item In modified ONS, We use the element wise multiplication denote by $\bigodot$ in Table \ref{tab:tab1} for reduction of matrices multiplication complexity.
\end{itemize}
To make a fair comparison, our modified ONS adopts the same experimental setting. We select randomly each feature vector $x_t$ at a time $t$, the regularization parameter $\lambda= 0.001$ and the learning rate $\eta_t= \frac{1}{2 \sqrt{t}}$. After that, all experimental results are reported by averaging the 20 runs with the help of Libol \cite{LIBOL2014}. We add our proposed algorithm in Libol library and compare all the state-of-art second order algorithms. \\
We compare the proposed second order modified Newton step algorithm with the existing second order online learning algorithms \cite{CW2008}, \cite{AROW2009}, \cite{SCW2016}. In addition ONS is the baseline technique with full information feedback. To accelerate the computational efficiency, the modify newton step employ a pure newton formula.
\begin{equation}
w_{t+1}=w_t - \eta_t \cdot A_t^{-1}b_t^T
\end{equation}
To compute $A_t^{-1}$ in time $\mathcal{O}(md)$ where $m$ is the number of classes in multiclass dataset and $d$ is feature vector in each record $t$, using the matrix-vector and vector-vector product after the $A_t^{-1}$ and $g_t^T$. \\
We have compared our algorithm against state-of-art algorithms using libol library \cite{LIBOL2014}.
We evaluate different second order online learning methods in the term of classification task. we select the Hinge loss $l(w_t)= max(0,1-y_tw_t^Tx_t)$ as a classifier. In this experiment, the step size of ONS $\eta_t=\frac{1}{2 \sqrt{t}}$. The results are shown in Figure 3. We find that the modify Newton step is much more efficient in term of running time and memory consumption. \\
In addition, the proposed second order online learning algorithm ONS converges faster than the other second order online learning methods used in experimentation. Hence, present approach provide a principled way to deal with large-scale online learning problems. Experimental results are presented using tables comparing the mistake rate, number of updates and time.\\
The Table \ref{Tab:1} is shows the summarized detail of datasets.

\begin{table}[h!]

\begin{center}

\begin{tabular}{|l|l|l|l|l|}
\hline

Dataset & $n$=\#instaces & $d$= \#features & $m=$ \#classes & type \\
\hline
satimage & 3104 & 36 & 6 & classification \\
\hline
mnist & 60000 & 780 & 10 & classification \\
\hline
acoustic & 78823 & 50 & 3 & classification \\
\hline
Protein & 17766 & 357 & 3 & classification \\
\hline
\end{tabular}

\end{center}
\caption{Summary of datasets in the experiment} \label{Tab:1}
\end{table}
The Table \ref{Tab:3} shows the empirical results of the satimage dataset, which contain 3104 instances, each instance contain 34 features and 6 classes. In table \ref{Tab:3}, columns contains the algorithm name, mistake rate, number of updates, and time in seconds for respective algorithm. Our approach performed better in time cost as compared other methods. For mistake rate, we are close to SCW and SCW2 and outperformed the CW and ARROW algorithm.

\begin{table}[h!]
\centering
	
	\begin{tabular}{|l|l|l|l|}

	\hline	
		Data Set:  & \multicolumn{3}{l|}{satimage  (\#instances=3104,\#features=36,\#classes=6)}                                                                                                                       \\ \hline
		Algorithm: & mistake        & \#updates                                                   & time (s)                                                                                \\ \hline
		CW    & 0.191 +/-0.004	 & 907.2 +/-21.0	&0.253 +/-0.020 \\ \hline
AROW       & 0.172 +/-0.008	 & 2106.6 +/-43.4	&0.308 +/-0.019  \\ \hline
		SCW1       & 0.153 +/-0.004	 & 795.2 +/-35.3	&0.265 +/-0.023   \\  \hline
		SCW2       & 0.157 +/-0.003	 & 1004.8 +/-35.7	&0.268 +/-0.032  \\ \hline
		ONS        & 0.171 +/-0.003	 & 1321.2 +/-19.8	& \textbf{0.171 +/-0.006}  \\   \hline
	\end{tabular}

\caption{ Results of satimage dataset} \label{Tab:3}

\end{table}

The Table \ref{Tab:4} present the experimentation results of the MNIST dataset, which contain 60000 instances and each instance contain 780 features with 10 classes. The MNIST dataset results, we are 40 times faster than other algorithms using the proposed algorithm. Our approaches also outperformed in the mistake rate which is less than in all second order online learning algorithm.

\begin{table}[h!]
\centering
	
	\begin{tabular}{|l|l|l|l|}

		\hline
		Data Set:  & \multicolumn{3}{l|}{mnist (\#instances=60000,\#features=780,\#classes=10)}                                                                  \\ \hline
		Algorithm: & mistake        & \#updates                                                   & time (s)                                                                                                      \\ \hline
		CW     & 0.133 +/-0.002	 & 15310.4 +/-87.9	&462.148 +/-2.911 \\ \hline
AROW       & 0.446 +/-0.079	 & 29992.8 +/-3590.6	&671.282 +/-80.369 \\ \hline
		SCW1       & 0.188 +/-0.002	 & 13809.0 +/-124.5	&313.469 +/-3.043  \\  \hline
		SCW2        & 0.130 +/-0.002	 & 15207.0 +/-56.1	& 344.911 +/-2.255 \\ \hline
		ONS        & \textbf{0.111 +/-0.001}	 & 21063.4 +/-124.9	& \textbf{5.902 +/-0.015} \\   \hline
	\end{tabular}

\caption{Results of MNIST  dataset} \label{Tab:4}
\end{table}

The Table \ref{Tab:5} shows the empirical results of the acoustic dataset, which contain the 78823 instances, 50 feature of each instance and 3 classes. our approach also outperformed in time as shown in the table \ref{Tab:5} and we find that our approach results in minimum updates.

\begin{table}[h!]

\centering
		\begin{tabular}{|l|l|l|l|}
\hline
		
			Data Set:  & \multicolumn{3}{l|}{acoustic  (\#instances=78823,\#features=50, \#classes=3)}                                               \\ \hline
			Algorithm: & mistake        & updates                                                   & time (s)                                                   \\ \hline
			CW    & 0.413 +/-0.001	 & 48342.6 +/-106.9	&9.224 +/-0.255  \\ \hline
			AROW       & 0.321 +/-0.001	 & 77391.6 +/-62.3	              &11.769 +/-0.338  \\ \hline
			SCW1       & 0.344 +/-0.009	 & 35305.4 +/-2004.5	&8.217 +/-0.392   \\  \hline
			SCW2       & 0.322 +/-0.001	 & 69253.4 +/-120.4	&11.250 +/-0.296  \\ \hline
			ONS        & 0.343 +/-0.007	 & \textbf{31932.2 +/-114.6}	& \textbf{4.213 +/-0.118}   	  \\   \hline
		\end{tabular}
	
	\caption{Results of acoustic  dataset} \label{Tab:5}
\end{table}

The table \ref{Tab:6} shows the results of the Protein dataset, which contains the 17766 instances, each instance contain 357 features and 3 classes. we also outperformed in the terms of time comparison and close to mistake rate achieved by other algorithms.

\begin{figure}[htb]

\caption{SATIMAGE dataset results} \label{Fig:1}
\includegraphics[width=1\textwidth]{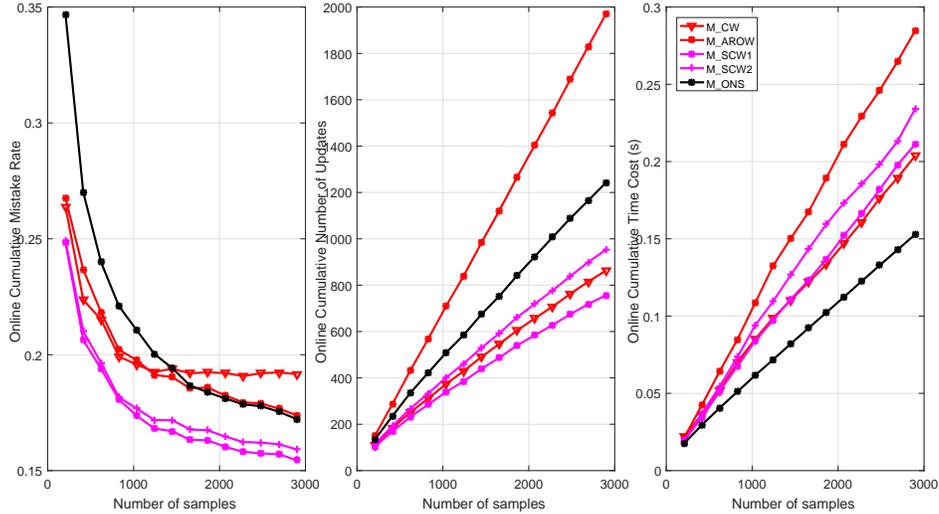}
\end{figure}

\begin{figure}[htb]
\caption{MNIST dataset results}
\label{Fig:2}
\includegraphics[width=1\textwidth]{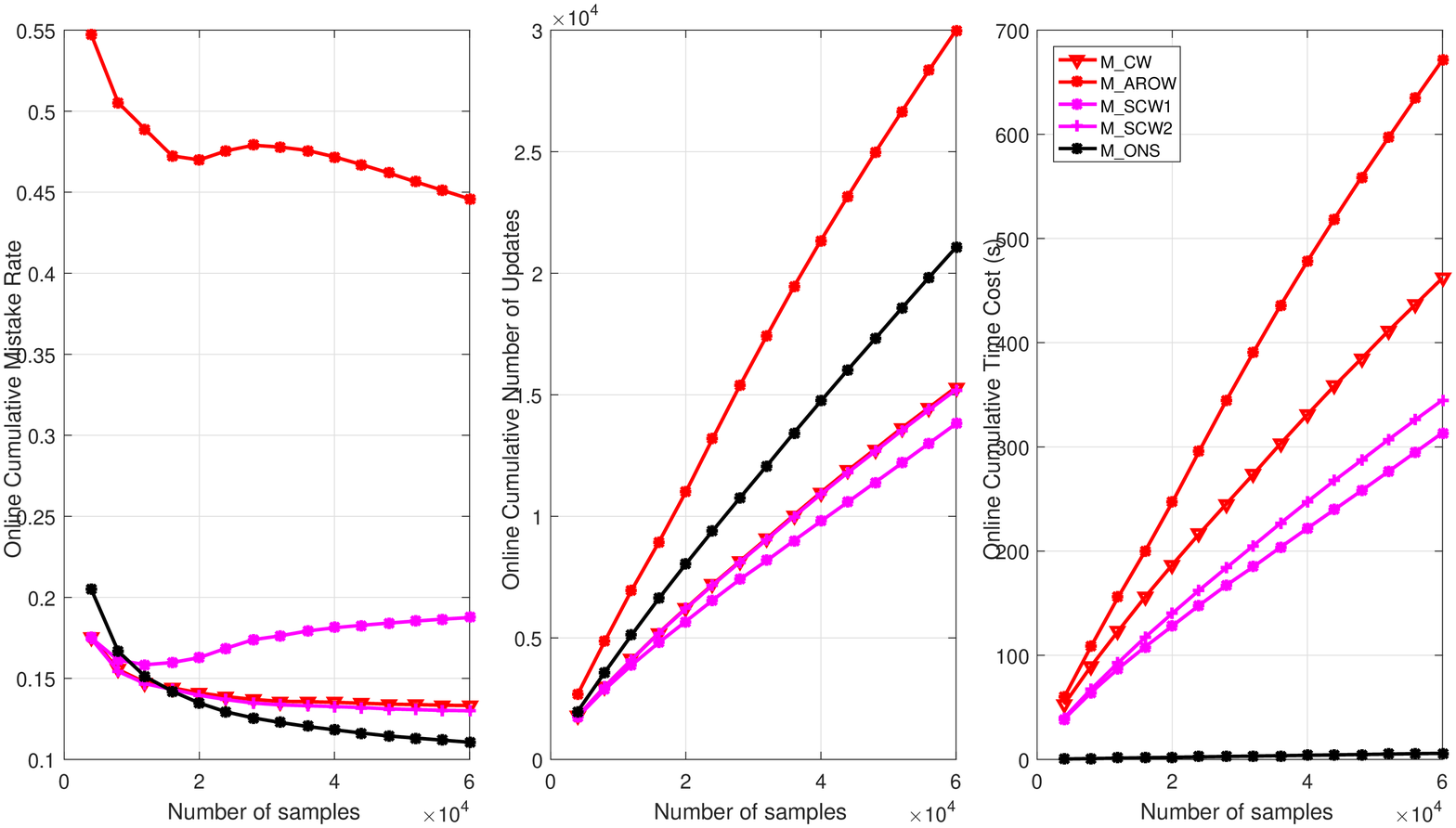}
\end{figure}

\begin{figure}[htb]

\caption{ACOUSTIC dataset results}
\label{Fig:3}
\includegraphics[width=1\textwidth]{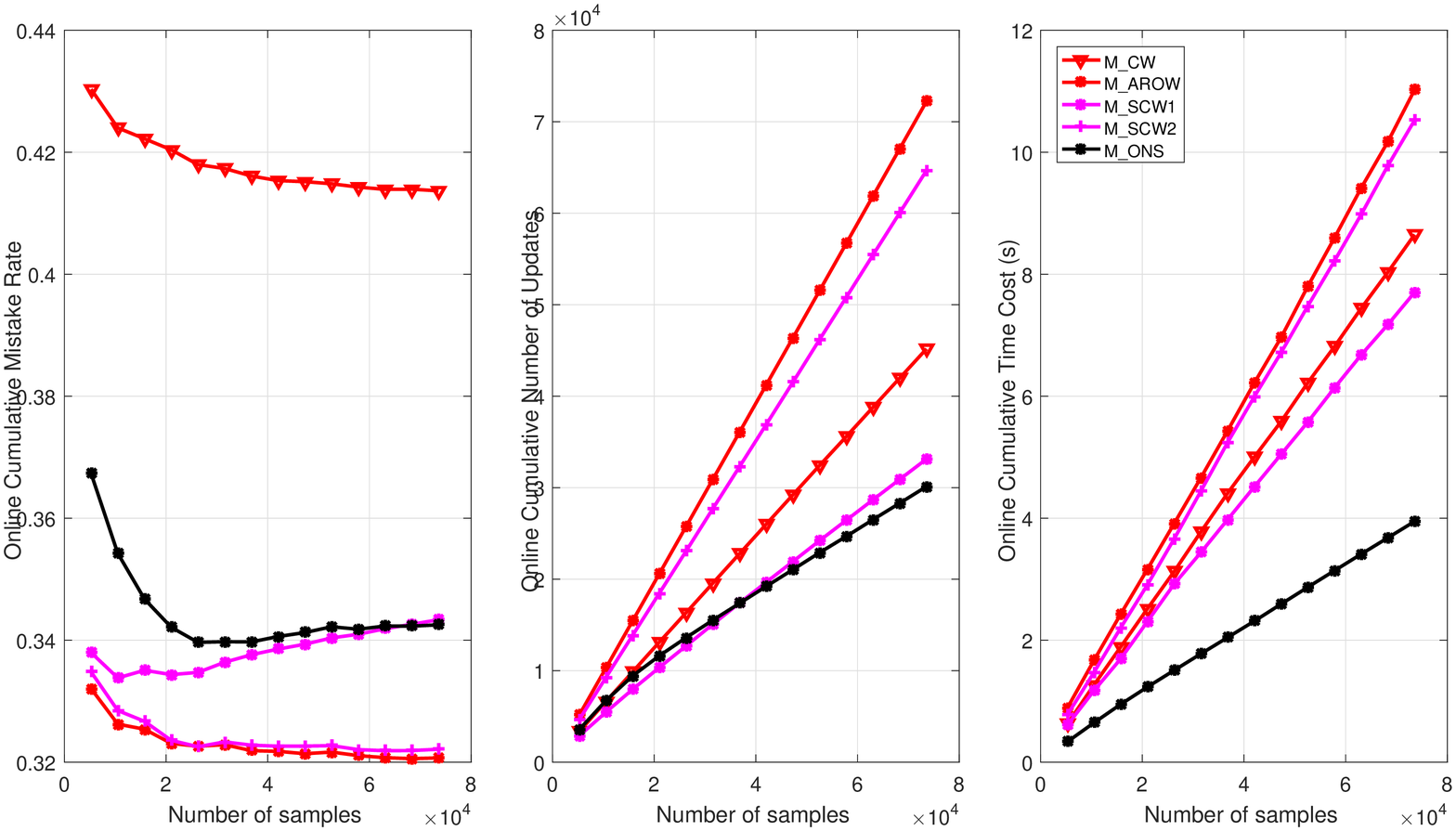}
\end{figure}

\begin{figure}[htb]

\caption{Protein dataset results}
\label{Fig:4}
\includegraphics[width=1\textwidth]{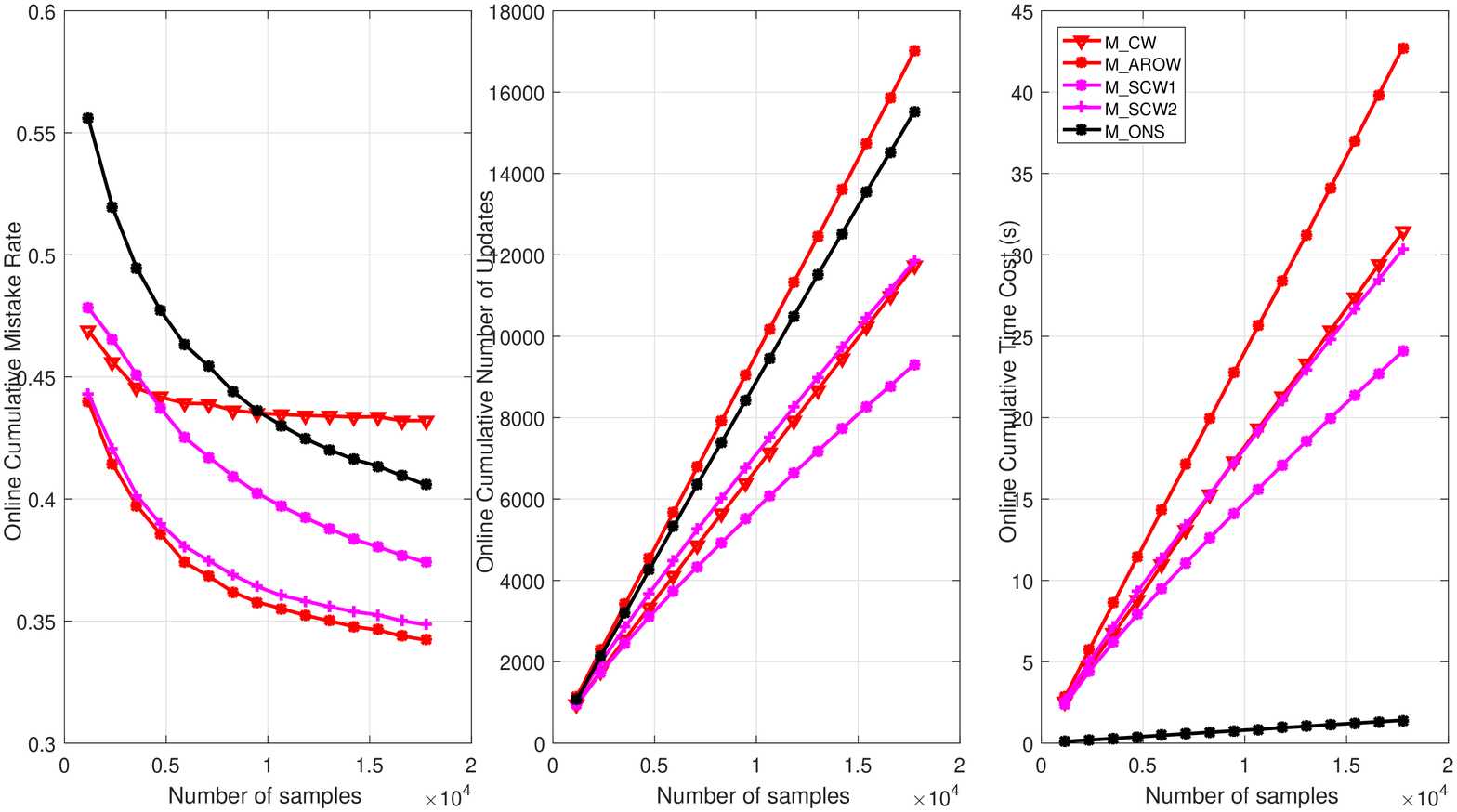}

\end{figure}

\begin{table}[h!]
\centering
	
		\begin{tabular}{|l|l|l|l|}
\hline
			
			Data Set:                                                             & \multicolumn{3}{l|}{Protein ( \#instances=17766, \#features=357, \#classes=3	 )}                                                                  \\ \hline
			Algorithm: & mistake        & updates                                                   & time (s)                                                                                                       \\ \hline
			CW     & 0.432 +/-0.002	 & 11742.4 +/-47.5	&31.420 +/-3.337 \\ \hline
			AROW        & 0.342 +/-0.002	 & 17012.4 +/-10.2	&42.742 +/-0.214 \\ \hline
			SCW1        & 0.374 +/-0.003	 & 9312.2 +/-42.6	&24.088 +/-0.115  \\  \hline
			SCW2      & 0.348 +/-0.001	 & 11848.4 +/-70.9	&30.351 +/-0.219 \\ \hline
			ONS        & 0.406 +/-0.002	 & 15525.0 +/-63.3	& \textbf{1.406 +/-0.006 }	  \\   \hline
		\end{tabular}
	
	\caption{Results of Protein  dataset} \label{Tab:6}
\end{table}
The figures [\ref{Fig:1}],[\ref{Fig:2}],[\ref{Fig:3}],[\ref{Fig:4}] shows the comparison of CW, AROW, SCW and SCW2 algorithms. Our proposed modified ONS shown in solid black line in figures [\ref{Fig:1}],[\ref{Fig:2}],[\ref{Fig:3}],[\ref{Fig:4}]. All figure include three parts, first part of figure shows the cumulative mistake rate, second part of fighure shows the number of updates and third part of figure shows the times in second. \\
The figure \ref{Fig:1} shows the comparison between second order online learning algorithms discuss in literature with the multiclass classification dataset satimage. We outperformed in terms of time taken in comparison to other methods and better than CW and AROW for the mistake rate.

The figure \ref{Fig:2} shows the results of MNIST dataset, our results are better in mistake rate and time comparison. The mistake rate is also less and time is 45 times less then other algorithms used in experimentation.
The figure \ref{Fig:3} shows the results of acoustic data, in which, we performed better in time.
The figure \ref{Fig:4} shown the result of Protein dataset, in which, we are 30 times faster than other algorithms.
\section{Conclusion}
The proposed method is an effort for second order online learning technique in order to address the challenging second order matrix size in large-scale dataset. 
We tackle the problem of second order matrix size by reducing the size of second order and first order matrices as $m \times d$, using element wise multiplication 
to retain the matrix size. We theoretically analyzed the regret bound of the proposed method and conducted a set of experiments to examine it's  empirical 
evaluation in training time and accuracy. We also applied large-scale classification datasets and results shows that our proposed algorithm do not need special 
hardware and could perform in common desktop machines. Our approach results in promising online computation time cost as discussed in experiment section. 
This further strengthen our approach as suitable second order learning to use in real time applications.







\end{document}